\RequirePackage{fix-cm}
\documentclass[smallextended]{svjour3}       % onecolumn (second format)
\smartqed  % flush right qed marks, e.g. at end of proof
\usepackage{graphicx}
\usepackage{amsmath, amssymb}
\usepackage{mathptmx}      % use Times fonts if available on your TeX system

\usepackage{enumerate}
\usepackage[misc]{ifsym}
\usepackage{float}
\usepackage[authoryear]{natbib}
\bibpunct{(}{)}{}{a}{}{;}

\begin{document}

\title{Generating Geological Facies Models with Fidelity to Diversity and Statistics of Training Images using Improved Generative Adversarial Networks}
%\thanks{Grants or other notes
%about the article that should go on the front page should be
%placed here. General acknowledgments should be placed at the end of the article.}

\author{Lingchen Zhu \and 
            Tuanfeng Zhang}
\institute{Lingchen Zhu (\Letter) \at
              Schlumberger-Doll Research, Cambridge, MA, USA\\  
              \email{Lzhu14@slb.com} \and Tuanfeng Zhang (\Letter) \at
              Schlumberger-Doll Research, Cambridge, MA, USA\\  
              \email{tzhang2@slb.com}}

%\date{Received: date / Accepted: date}
\maketitle

\begin{abstract}
This paper presents a methodology and workflow that overcome the limitations of the conventional Generative Adversarial Networks (GANs) for geological facies modeling. It attempts to improve the training stability and guarantee the diversity of the generated geology through interpretable latent vectors. The resulting samples are ensured to have the equal probability (or an unbiased distribution) as from the training dataset. This is critical when applying GANs to generate unbiased and representative geological models that can be further used to facilitate objective uncertainty evaluation and optimal decision-making in oil field exploration and development. 

We proposed and implemented a new variant of GANs called “Info-WGAN” for the geological facies modeling that combines Information Maximizing Generative Adversarial Network (InfoGAN) with Wasserstein distance and Gradient Penalty (GP) for learning interpretable latent codes as well as generating stable and unbiased distribution from the training data. Different from the original GAN design, InfoGAN can use the training images with full, partial, or no labels to perform disentanglement of the complex sedimentary types exhibited in the training dataset to achieve the variety and diversity of the generated samples. This is accomplished by adding additional categorical variables that provide disentangled semantic representations besides the mere randomized latent vector used in the original GANs. By such means, a regularization term is used to maximize the mutual information between such latent categorical codes and the generated geological facies in the loss function. 

Furthermore, the resulting unbiased sampling by Info-WGAN makes the data conditioning much easier than the conventional GANs in geological modeling because of the variety and diversity as well as the equal probability of the unconditional sampling by the generator.

% \PACS{PACS code1 \and PACS code2 \and more}
% \subclass{MSC code1 \and MSC code2 \and more}
\end{abstract}

%%%%%%%%%%%%%%%%%%%%%%%%%%%%%%%%%%%%%
%%%%%%%%%%%%%%%%%%%%%%%%%%%%%%%%%%%%%
\section{Introduction and methodology}
\label{sec:1}
\begin{figure}[H]
\begin{center}
\includegraphics[width=1.0\textwidth]{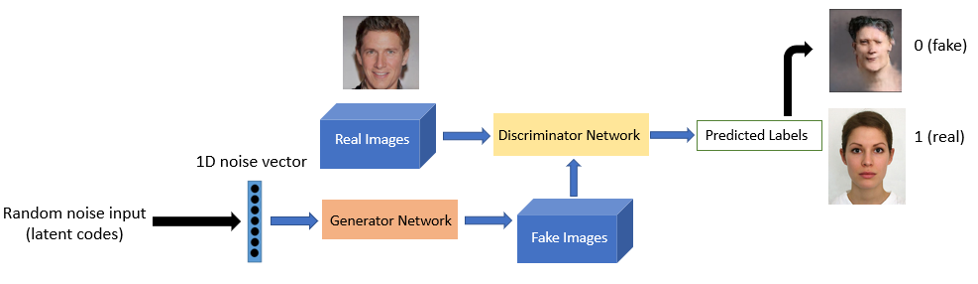}
\end{center}
\caption{Schematic diagram showing the structure of GANs that consists of two networks: one generator and another for discriminator. The generator generates fake images to fool the discriminator while the discriminator aims to distinguish the real images from the training data from the fake ones by the generator.}
\label{fig:gandiagram}   
\end{figure}

Generative Adversarial Networks (GANs) \citep{goodfellow2014generative} are gaining popularity in the deep learning industry, particularly in computer vision by generating photo-realistic images (\cite{johnson2016super}; \cite{zhu2018cyclegan}; \cite{karras2018progressivegan}). GAN is a generative model composed of a generator and a discriminator, each parametrized by a separate neural network. The generator is trained to map a latent vector into an image, while the discriminator is trained to distinguish the real (training) images from those that have been generated by the generator. The goal of GANs is to train both the generator and the discriminator such that the generator can create images that the discriminator cannot tell the difference.  Both the generator and the discriminator are considered as two players that play a minimax game in an adversarial fashion with the following loss function:

\begin{equation}
L =\min_{G} \max_{D}V(D, G)=\mathbb{E}_{\mathbf{x} \sim p_{data}} [\log D(\mathbf{x})] + \mathbb{E}_{\mathbf{z} \sim p_{noise}} [\log (1 - D(G(\mathbf{z})))]
\label{eq:1}
\end{equation}

where $V(D, G)$ represents the reward (a.k.a., loss) for the discriminator that aims to maximize its value by forcing $D(\mathbf{x})$ to approach 1 and $D(G(\mathbf{z}))$ being as close as 0 while the generator tends to minimize its loss by boosting $D(G(\mathbf{z}))$ to be 1. See Figure~\ref{fig:gandiagram} for a schematic illustration of the GANs.

A more efficient and practical structure of GANs was proposed by Radford et al \citep{radford2015unsupervised}, which introduced deep convolutional generative adversarial networks (DCGAN) to learn a hierarchy of representations from object parts to scenes in both the generator and discriminator. This DCGAN structure has become a standard implementation of GANs in general image representations and image generations.

The emerging application areas of GANs include physics, astronomy, chemistry, biology, health care, geology, arts and others. Despite these promising applications, training GANs is notoriously difficult because of a well-known phenomenon called “mode collapse” that the generator produces very limited varieties of samples, causing either non-converged or vanishing gradients in the process of GAN training. The biggest disadvantage resulted from the mode collapse is the biased sampling in GANs that tends to compromise the use of GAN generated samples to make predictions when the uncertainty needs to be considered and addressed in an objective manner. This is particularly true when applying GANs to model geology.

Another disadvantage of GAN is that the latent vector used for image generation is highly entangled, i.e., one cannot know if each separate element in the latent vector could have any semantic meaning. Therefore, the latent vector lacks the ability to interpret the salient attributes of the data, like fluvial, deltaic, etc., in our application of geological facies modeling.

The pathological mode collapse phenomenon in the original GAN approach was later discussed in a publication by Chen et al. \citep{chen2016infogan} and was explained because of the entangled information embedded in the latent space. They proposed an extension to the original GANs such that it can learn disentangled representation in a completely unsupervised or semi-supervised manner. The authors introduced additional latent codes $\mathbf{c}$ on top of a simple continuous input noise vector $\mathbf{z}$ as illustrated in Figure~\ref{fig:gandiagram} to impose selective representations in a disentangled manner to overcome the limitation in the generator since it creates samples in a highly entangled way due to the lack of the correspondence between the individual dimensions of $\mathbf{z}$ and the semantic features of the data.
   
This approach was named as InfoGAN \citep{chen2016infogan} and it can generate samples with the variety in the training data by maximizing the discerning capability of each code, or label to its associated images using the mutual information concept in information theory. The demonstrative examples include the disentangling of the writing styles from digit shapes on the MNIST dataset, pose from lighting of 3D rendered images and background digits from the central digit on the SVHN dataset.

\begin{figure}[H]
\begin{center}
\includegraphics[width=1.0\textwidth]{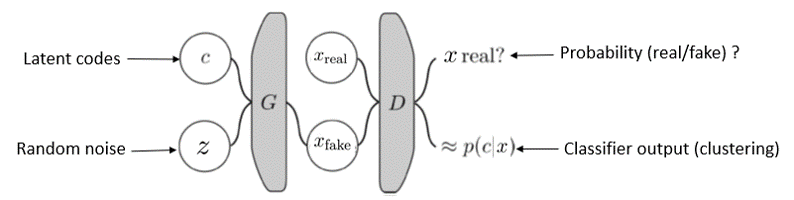}
\end{center}
\caption{Schematic diagram of InfoGAN structure.}
\label{fig:infogandiagram}   
\end{figure}

Figure~\ref{fig:infogandiagram} illustrates the InfoGAN structure that is like GAN structure in Figure~\ref{fig:gandiagram} except for the added latent codes $\mathbf{c}$ in the input layer and an extra classifier in the output layer of the network that provides the categorical probability $p(\mathbf{c}|\mathbf{x})$ given the input $\mathbf{x}$.

The InfoGAN loss is defined as the following:

%\begin{minipage}{5.5in} 
%\setlength{\mathindent}{0.0in}
\begin{equation}
V_{InfoGAN}(D, G)=\mathbb{E}_{\mathbf{x} \sim p_{data}} [\log D(\mathbf{x})] + \mathbb{E}_{\mathbf{z} \sim p_{G}} [\log (1 - D(G(\mathbf{z})))] - \lambda I(\mathbf{c}; G(\mathbf{z}, \mathbf{c}))
\label{eq:2}
\end{equation}
%\end{minipage} 

In addition to InfoGAN, another improvement on the training stability and eliminating the mode collapse of the original GAN was presented by Arjovsky et al. \citep{arjovsky2017wgan}. This paper introduced a new approach, named Wasserstein GAN (WGAN) with gradient penalty, to stabilize the GAN training. WGAN uses an earth-mover (Wasserstein) distance to achieve a much smoother loss function that reduces the possibility of GANs getting stuck in local minimums, which is highly likely in the original GAN and easily causes vanishing gradients for training. The Wasserstein distance is defined as:

\begin{equation}
W(P_{data}, P_{G})=\inf_{\gamma \sim \Pi (P_{data}, P_{G})}\mathbb{E}_{(\mathbf{x}, \mathbf{y}) \sim \gamma} [\|  \mathbf{x}-\mathbf{y} \|]=\sup_{{\| f \|}_{L}} \mathbb{E}_{\mathbf{x} \sim p_{data}}[f_{\mathbf{w}}(\mathbf{x})] - \mathbb{E}_{\mathbf{x} \sim p_{G}}[f(\mathbf{x})]
\label{eq:3}
\end{equation}

where the function $W(P_{data}, P_{G})$ is the earth-mover distance that is formally defined as the minimum cost of transporting mass in order to transform the real data distribution $P_{data}$ to the generated data distribution $P_{G}$, and $f$ is an arbitrary function defined in the real field. In our context, $f_{\mathbf{w}}$ is the discriminator neural network with weights $\mathbf{w}$.

On top of Wasserstein distance, gradient penalty is applied to further boost the training stability of GANs, which penalizes the gradient whose norm is away from one by the following updated loss function (\cite{gulrajani2017gp}; \cite{wei2018gp}):

\begin{equation}
L_{(WGAN-GP)}W(P_{data}, P_{G})=-W(p_{data}, p_{G}) + \lambda \mathbb{E}_{x \sim P_{data}}(\| {\nabla}{f_{w}(\mathbf{x})} \| - 1)^2
\label{eq:4}
\end{equation}

The finalized approach that combines InfoGAN and WGAN with gradient penalty is denoted as Info-WGAN-GP and is referred as Info-WGAN for brevity in this paper.

Recently, researchers started applying the original GANs to geology and reservoir engineering. Specifically, Chan and Elsheikh \citep{chan2017parameterizing} proposed parameterizing a geological model using GANs. GANs have also been proposed to reconstruct porous medium from CT-scan rock samples \citep{mosser2017ctscan}. The use of GANs in geostatistical inversion was discussed by Laloy et al. \citep{laloy2017ganinversion}.

The paper by Dupont et al. \citep{dupont2018gangeomodeling} was the first publication using GANs to generate geological models at the reservoir scale constrained to well data. A library of reservoir-scale 2D models was generated by object-based modeling (abbreviated as OBM) (\cite{holden1998obm}; \cite{skorstad1999obm}) as training images that exhibit and represent a wide variation of depositional facies patterns. A semantic inpainting scheme (\cite{li2017inpainting}, \cite{pathak2016inpainting}, \cite{yeh2016inpainting}) was used to generate conditional models by GANs that fully honor well data. 

Later, an extension of the work by Dupont et al. to 3D was presented by Zhang et al. \citep{zhang2019gan3dmodeling}. It has been demonstrated that the 3D GANs outperforms the advanced geostatistical reservoir modeling approaches such as multi-point statistics (MPS) in generating more geologically realistic facies models constrained by well data, particularly when the subsurface geology contains non-stationary and heterogeneous geological sedimentary patterns such as progradational and aggradational trend, which is a ubiquitous phenomenon in most reservoirs.

Despite the promising applications of GANs in geological modeling, we have observed several key issues that would prevent its successful use in building faithful reservoir facies models that are necessary for the objective uncertainty evaluation and optimal decision-making in exploration and field developments in the oil industry. The root cause of these issues is the frequent mode collapse in the training of the original GAN method, leading to severely biased samples by the generator and the lack of diversity in the resulting models, which further compromise the usefulness of the facies models by GANs.

This paper aims to apply the Info-WGAN as a novel tool in modeling geology by eliminating pathological mode collapse phenomenon in the original GAN method to provide efficient solutions in building geological facies models.  The resulting facies models by Info-WGAN have the exact diversity and equal probability inherited from the training dataset and provide interpretable attributes as a few disentangled elements in the latent variable space, to allow generating desireble geological facies models specified by the user. We believe that our novel workflow is beneficial not only in geology but also in other domains when applying GANs for modeling the respective phenomena.   

\section{Generating equal probable realizations of the geology using Info-WGAN}
Even though GANs can generate different geological facies models by learning the representation of the sedimentary facies associations from the training images, the samples that are created by GAN generator could be highly biased. This section will demonstrate how we resolve this major limitation in GANs using Info-WGAN.
\label{sec:2}
\subsection{Case 1: binary fluvial facies}
\begin{figure}[H]
\begin{center}
\includegraphics[width=0.95\textwidth]{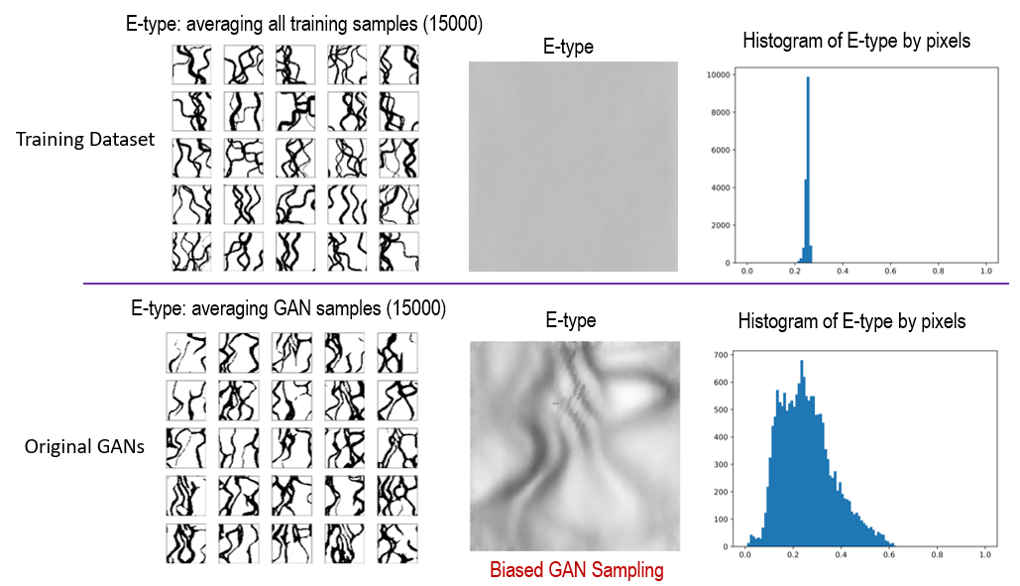}
\end{center}
\caption{Diversity and equal probability of sand distribution in the training data vs. the biased sampling from the conventional GANs}
\label{fig:ganbias}    
\end{figure}

Figure~\ref{fig:ganbias} shows some training examples from 15000 binary fluvial training images (left-most on the top of the figure with channel sand: black, and shale background: white). 

The main flow direction of the channels is from north to south with varying channel width, sinuosity and amplitudes. The channels are distributed evenly in space, i.e. they can happen anywhere in the 2D area that confines the channels in the training images. This can be verified by an e-type map (estimation type) of the training images. The e-type map is computed as an estimation of the channel sand probability in space by performing pixel-wise averaging of all channel images with channel sand being assigned to a value 1 and 0 for the background.

The e-type map in the middle of the top of Figure~\ref{fig:ganbias} is almost a flat (constant) map, which indicates the channels from the training dataset are evenly spaced and pixels in all the training images are considered as equal probable, meaning that the channels can be at any position with equal probability. This is also a basic assumption required by traditional geostatistical simulations using Monte Carlo sampling \citep{deutsch1998gslib}. The approximately constant e-type value is around the mean value of the channel sand proportion from 15000 training images, which can be manifested by a tight histogram of all the e-type pixel values with the mean value being equal to the sand proportion in each of the training images (=0.25 in this case study).

However, if the original GAN method is used, the trained generator becomes highly biased due to the mode collapse even though the samples reproduce the geometry of the channels from the training dataset reasonably (most-left at the bottom of Figure~\ref{fig:ganbias}). In contrast to the e-type of the channel training images that is constant and flat, the e-type of the generated samples by GANs (middle at the bottom of Figure~\ref{fig:ganbias}) shows that channel sand happens more likely in some area than others, indicating a highly biased learning of GANs. This also suggests that GANs fails to capture the true data distribution during the training, which is manifested in the histogram of e-type with a wide spread of pixel values from 0 to 0.6 (most-right at the bottom of Figure~\ref{fig:ganbias}).

This biased sampling is one big limitation when using GANs for geological modeling because all samples cannot be claimed as truly realistic realizations since they are not equally probable like we normally use in geostatistical simulation. Consequently, all the samples (static facies models) cannot be used for further uncertainty evaluation and propagation when the static models are fed into flow simulations. Moreover, the e-type of such samples becomes less meaningful since their distribution is different from that of the training dataset, and therefore the e-type cannot be treated to be sand probability map anymore.   

\begin{figure}[H]
\begin{center}
\includegraphics[width=1.0\textwidth]{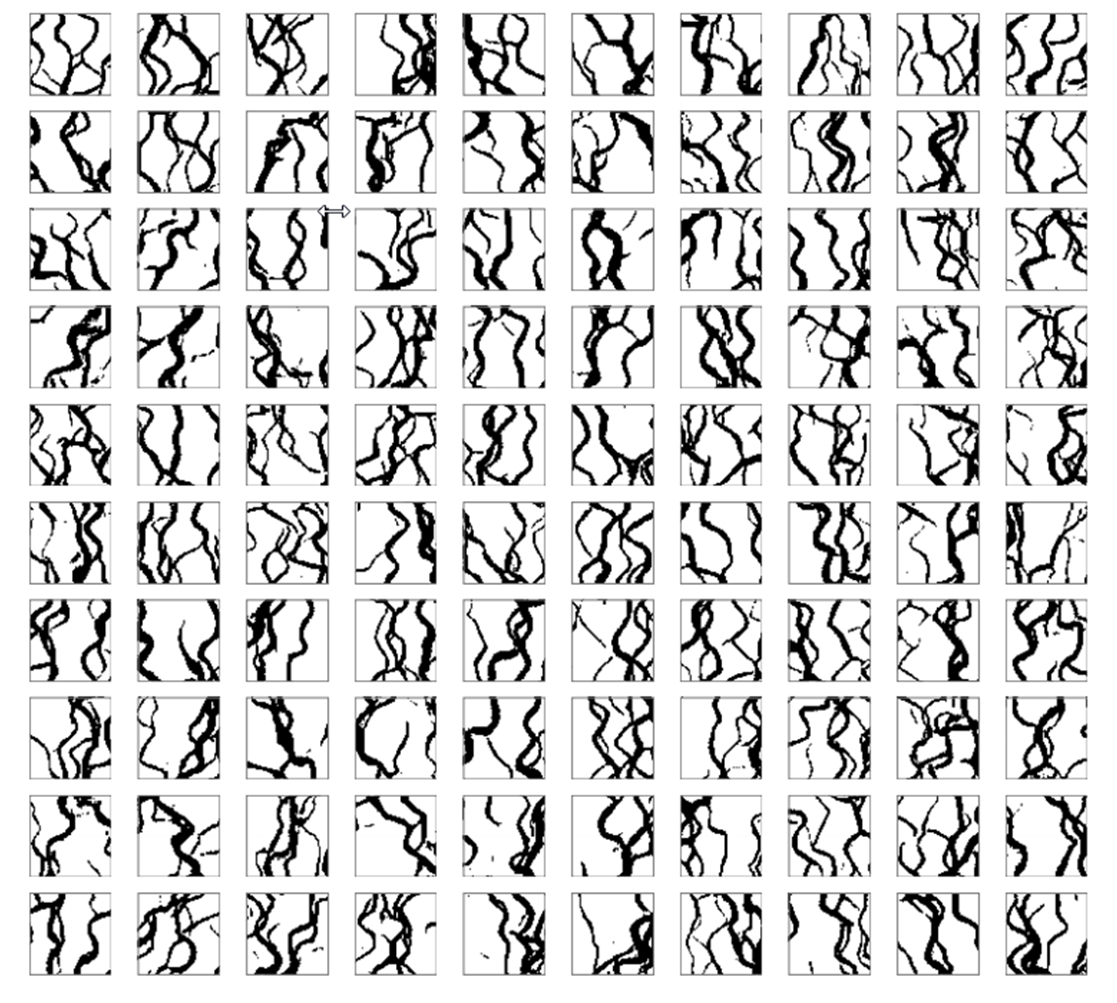}
\end{center}
\caption{Binary facies samples by Info-WGAN reproduces the channel geometry reasonably}
\label{fig:infogan100samplesbinaryfluvial}    
\end{figure}

To overcome this limitation, we applied Info-WGAN to generate facies samples. The same set of 15000 fluvial training dataset were used to train the model. We used one dummy categorical latent code that indicates all the channels to be the same type such that the infomation maximization component is temporarily disabled and only the Wasserstein distance and gradient penalty are utilized. Figure~\ref{fig:infogan100samplesbinaryfluvial} shows 100 samples by Info-WGAN that suggests reasonable reproduction of the geometry of fluvial deposits. 

\begin{figure}[H]
\begin{center}
\includegraphics[width=1.0\textwidth]{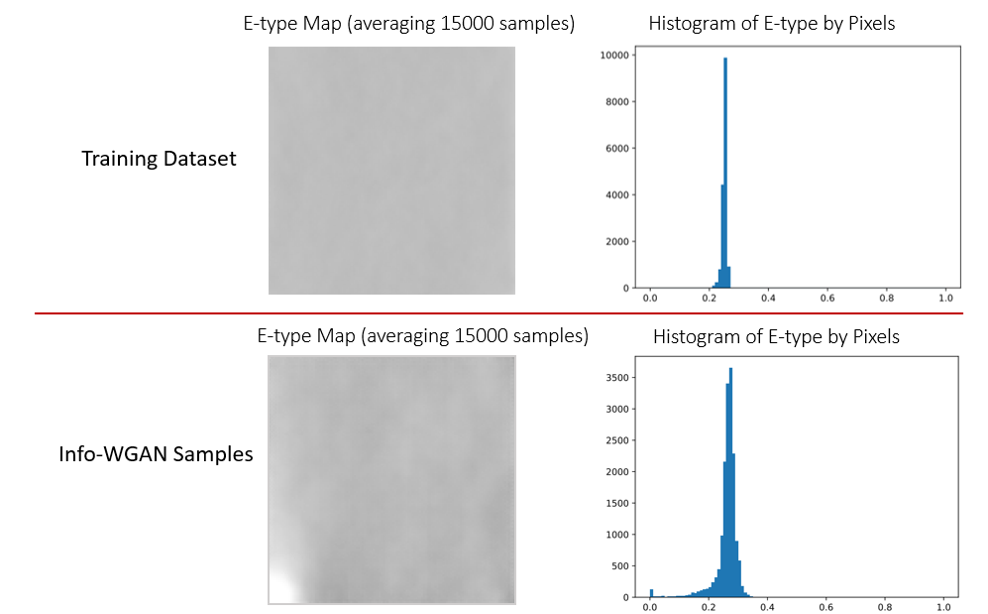}
\end{center}
\caption{Diversity and equal probable sand distribution in the training dataset (top) and they are reasonably reproduced by Info-WGAN (bottom)}
\label{fig:etypeinfoganbinaryfluvial}    
\end{figure}

Thanks to the Wasserstein distance and the gradient penalty used in the model, further verification on the e-type of 15000 samples suggests that Info-WGAN can generate diverse and equal probable realizations of channels by learning the true distribution of the training dataset, a striking contrast to the original GANs. This is manifested in the e-type map (most-left at the bottom of Figure~\ref{fig:etypeinfoganbinaryfluvial}) with the corresponding tight histogram of the pixel values, which are close to the statistics from the training dataset (the top of Figure~\ref{fig:etypeinfoganbinaryfluvial}). The e-type map by the InfoGAN is very close to that from the training images except for a small artifact area (bright spot) at the lower-left corner.

That Info-WGAN is much more superior than the original GANs in creating diverse as well as equal probable samples can be explained by its use of Wasserstein distance and gradient penalty to stabilize the GAN training to avoid mode collapse.

\subsection{Case 2: mixed 2 types of fluvial and deltaic systems}

In this case study, we move further to test the disentangling capability of Info-WGAN by mixing fluvial and deltaic channel training images to see whether Info-WGAN is capable to reproduce both types with equal probable realizations by reproducing the correct sand statistics from the training dataset.   

\begin{figure}
\begin{center}
\includegraphics[width=1.0\textwidth]{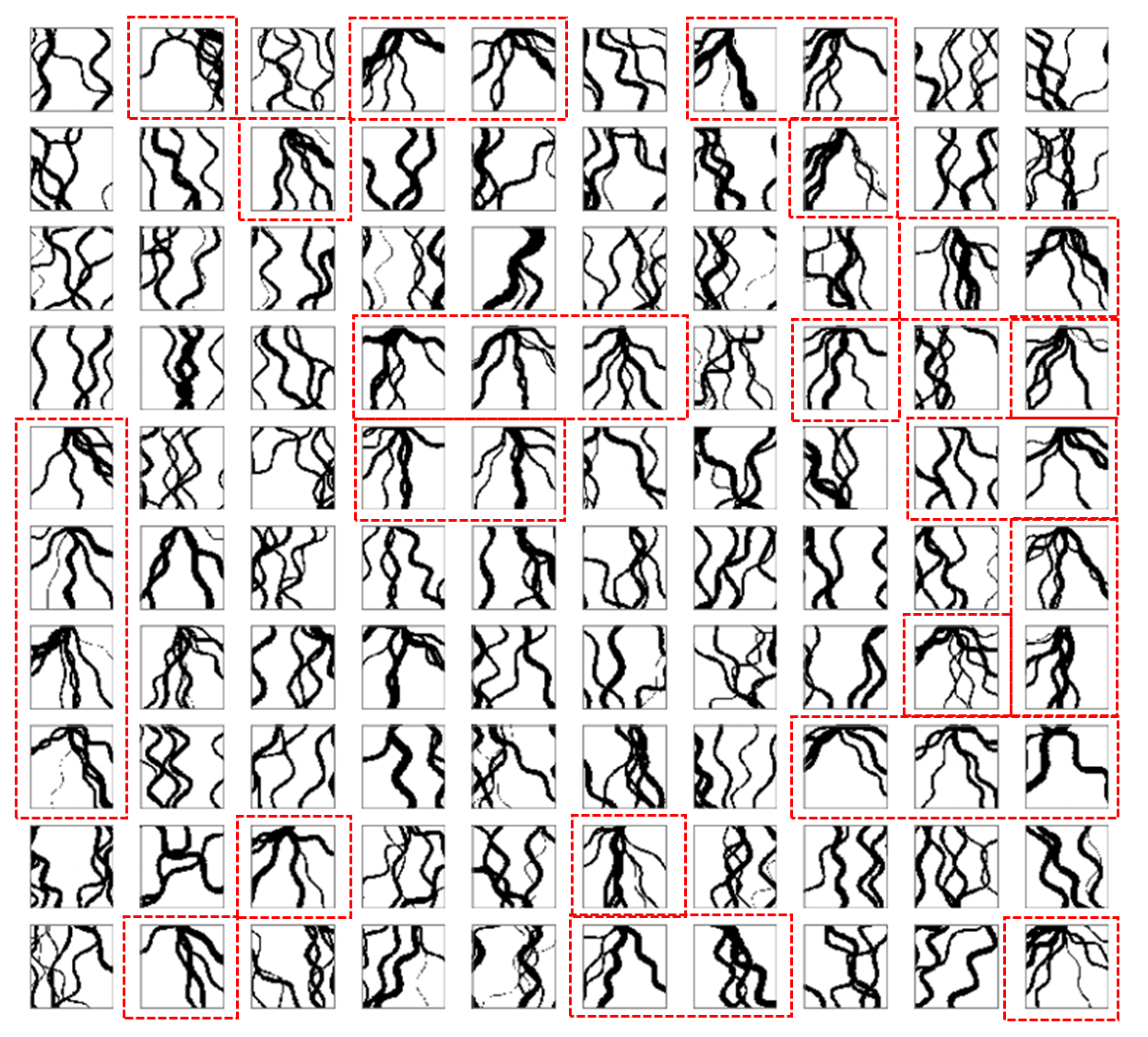}
\end{center}
\caption{100 of the total 15000 training images that contains mixed two types of deposits: one is fluvial (10000) and another 
for deltaic (5000) that is enclosed by dashed red squares}
\label{fig:training100samplesfluvialplusdeltaic}    
\end{figure}

Figure~\ref{fig:training100samplesfluvialplusdeltaic} shows 100 of the total 15000 training images that are generated by OBM, which is a mix of two types depositional environments: fluvial (type-I) and deltaic (type-II). The fluvial channels mainly follow the north-south direction while the deltaic system starts form a point source at the middle of the upper border of the region and then spreads out to the south direction. This dataset creates challenges for the original GANs to learn the diversity and generate two types of systems due to the mode collapse limitation.

\begin{figure}
\begin{center}
\includegraphics[width=1.0\textwidth]{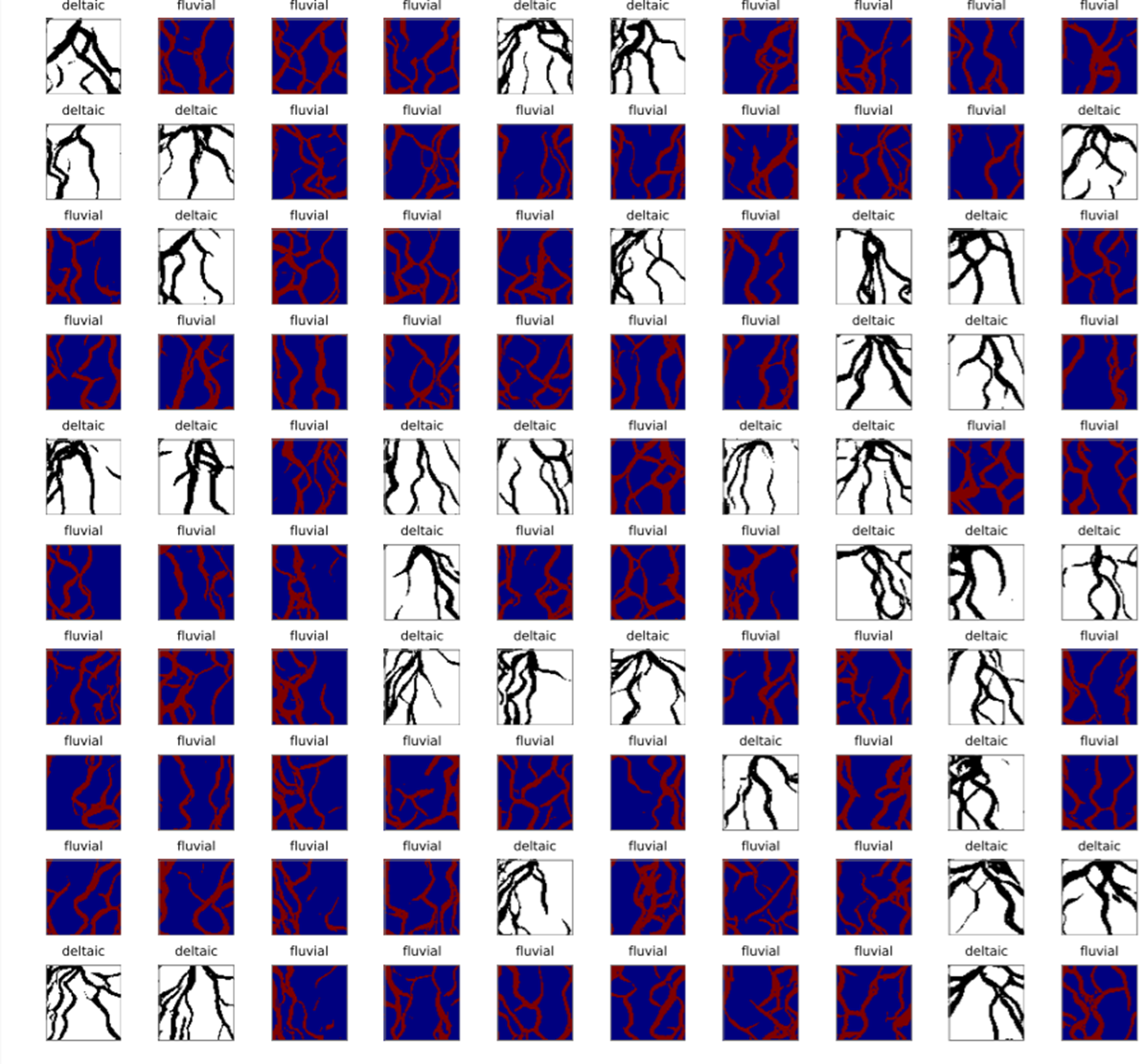}
\end{center}
\caption{100 samples by the Info-WGAN that reproduce the mix of two types of deposits reasonable with correct labels}
\label{fig:infogan100samplesfluvialplusdeltaic}    
\end{figure}

We used the Info-WGAN in a novel way by introducing two bits of categorical codes $\mathbf{c}$: $\mathbf{c}$ = [0, 1] for the fluvial deposits and $\mathbf{c}$ = [1, 0] for the deltaic deposits. This can be considered as the use of Info-WGAN in supervised fashion; however, the labels assigned to images by sedimentary types are easy to provide. After training Info-WGAN, the generator can create both systems with the correct labels. Figure~\ref{fig:infogan100samplesfluvialplusdeltaic} shows 100 samples by Info-WGAN that gives a satisfactory mix of the fluvial and deltaic deposits, a striking contrast with the original GANs that can generate either only the fluvial or deltaic deposits but not both due to the use of a single noisy vector in the latent space. Because of maximizing the mutual information carried out by the two additional categorical codes, Info-WGAN can generate a mix of the two different types of environments through disentangling capability without encountering the mode collapse issue that normally happens in the original GANs. Meanwhile, it is worth noting that the use of both Wasserstein distance and gradient penalty also contribute to the training stability of Info-WGAN. 

\newpage

\begin{figure}[H]
\begin{center}
\includegraphics[width=1.0\textwidth]{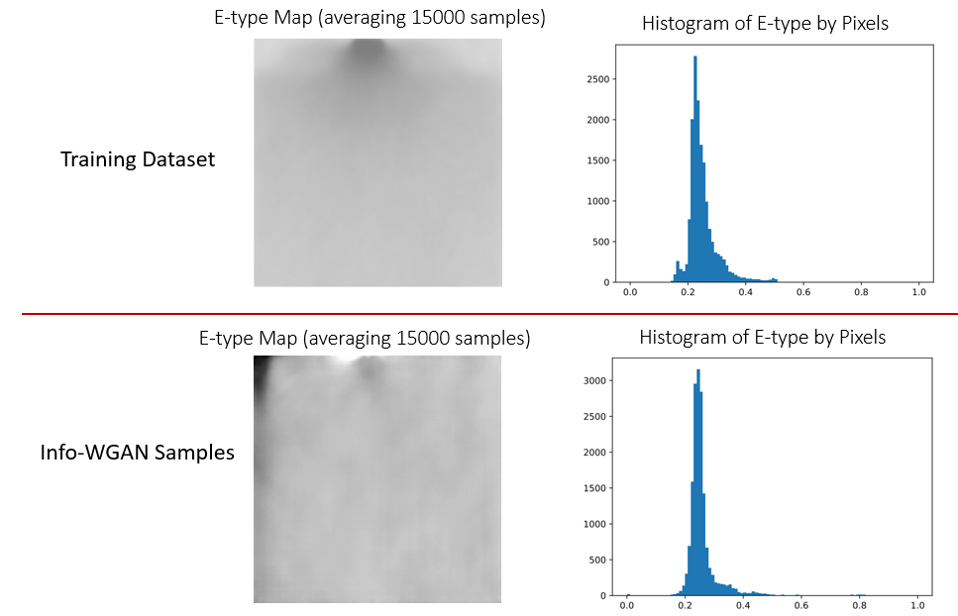}
\end{center}
\caption{E-type of the samples (bottom) demonstrates the reasonable production of the two types of deposits by Info-WGAN in terms of the diversity and statistics}
\label{fig:etypeinfoganbinaryfluvialplusdeltaic}    
\end{figure}

The e-type map of the sand facies and its statistics by pixels from 15000 samples by Info-WGAN are similar to those from the training images, which show a slightly darker regions at the top of the area because of the more concentrated channel sand in the deltaic system (see Figure~\ref{fig:etypeinfoganbinaryfluvialplusdeltaic}). This test case suggests that Info-WGAN can generate the mix of two types of sedimentary systems with equal probable realizations by reproducing the correct sand statistics from the training dataset.

\subsection{Case 3: deltaic system with 4 facies}

This case study demonstrates that the Info-WGAN can generate equal probable realizations by reproducing the correct statistics for each facies when there are multiple ($>2$) facies in the sedimentary system.  

\begin{figure}
\begin{center}
\includegraphics[width=0.90\textwidth]{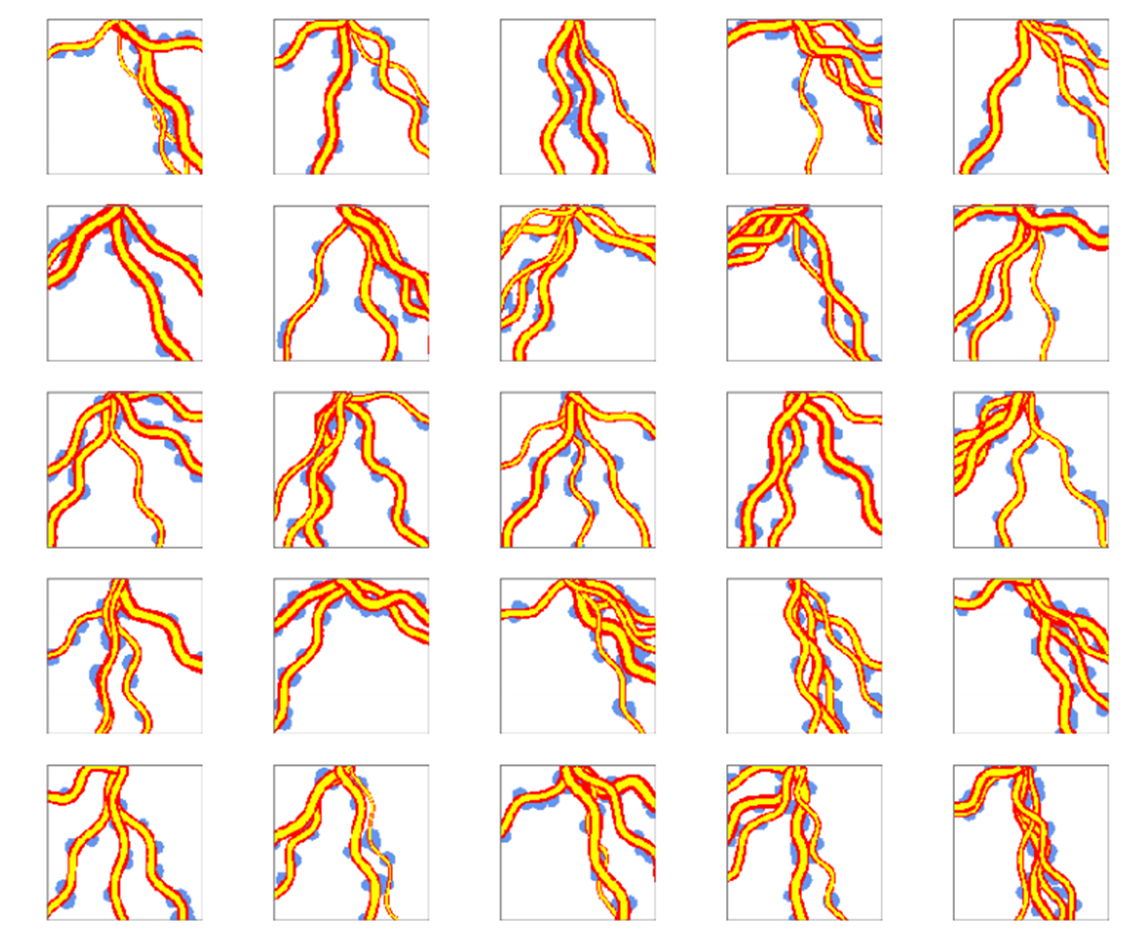}
\end{center}
\caption{25 of 10000 deltaic training images with 4 facies (channel: yellow, levee: red, splay: blue, shale: transparent)}
\label{fig:infogan25trainingdeltic4facies}    
\end{figure}

Figure~\ref{fig:infogan25trainingdeltic4facies} shows 25 of total 10000 deltaic training images with 4 facies: channel, levee, splay and shale background. The facies association can be clearly observed by the following relationships: the channel sand (yellow) is bounded by the levee (red) that attaches the splay (blue), which is embedded in the shale background (transparent).

\begin{figure}[H]
\begin{center}
\includegraphics[width=0.90\textwidth]{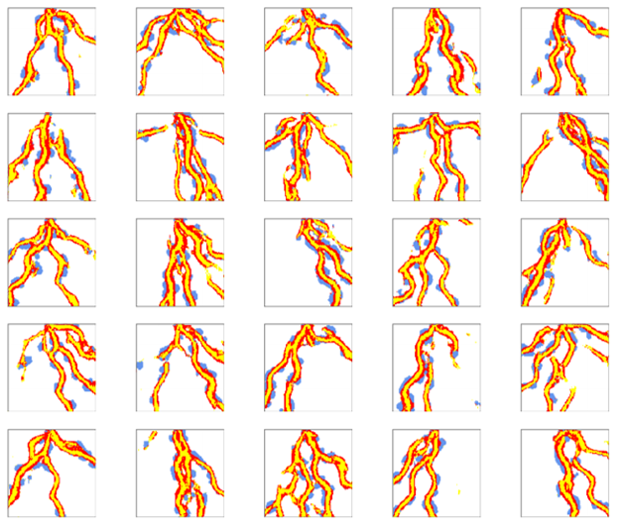}
\end{center}
\caption{25 deltaic samples by Info-WGAN with 4 facies that show reasonable reproduction of facies associations}
\label{fig:infogan25samplesdeltic4facies}    
\end{figure}

Info-WGAN is used with one constant categorical code to train the both generator and discriminator networks and then the trained generator is used to produce samples. Figure~\ref{fig:infogan25samplesdeltic4facies} shows 25 samples generated by the Info-WGAN that demonstrate reasonable reproduction of the facies geometric relationships, connectivity and their association. 

\begin{figure}[H]
\begin{center}
\includegraphics[width=1.0\textwidth]{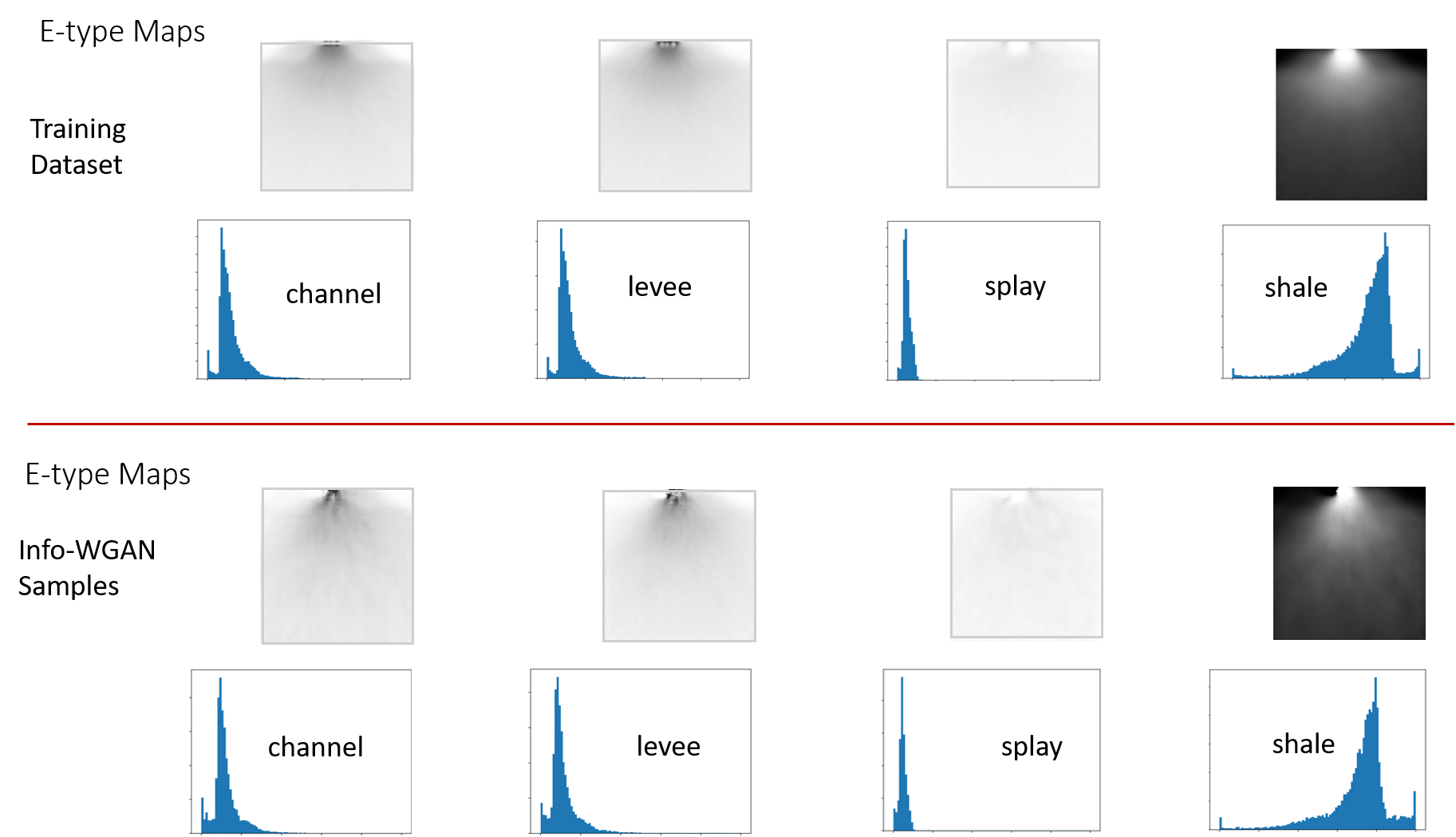}
\end{center}
\caption{The e-type maps and their histogram statistics for each of the 4 facies in the training images (top) are well reproduced by the samples generated by Info-WGAN (bottom). In each histogram, the x-axis has the rane [0, 1] and the y-axis is the freqency.}
\label{fig:etypeinfogandeltaic4facies}    
\end{figure}

Further testing of the e-type maps foe each individual facies is displayed in Figure~\ref{fig:etypeinfogandeltaic4facies}. When computing the e-type for one specified sedimentary facies, an indicator transformation is applied, i.e. , the corresponding studied facies is indicated as 1 and others as 0. The e-type map of a specified facies is then created by pixel-wise averaging of indicator maps. In Figure~\ref{fig:etypeinfogandeltaic4facies}, we can observe that the e-type maps and their statistics for all the facies from the training images are reproduced quite well by the generator of the Info-WGAN.

\newpage

\subsection{Case 4: mixed 3 types of systems with different number of facies}

This case study pushes the envelope to test the boundaries of the applicability of Info-WGAN. We merged all three types of sedimentary systems discussed above into one rich training dataset that contains 5000 binary fluvial images, 5000 binary deltaic images (called deltaic-I) and 5000 additional deltaic images with 4 facies (called deltaic-II). The facies coding is consistent in the mixed training images as 1 for channel, 2 for levee, 4 for splay and 0 for the background shale.

\begin{figure}[H]
\begin{center}
\includegraphics[width=0.925\textwidth]{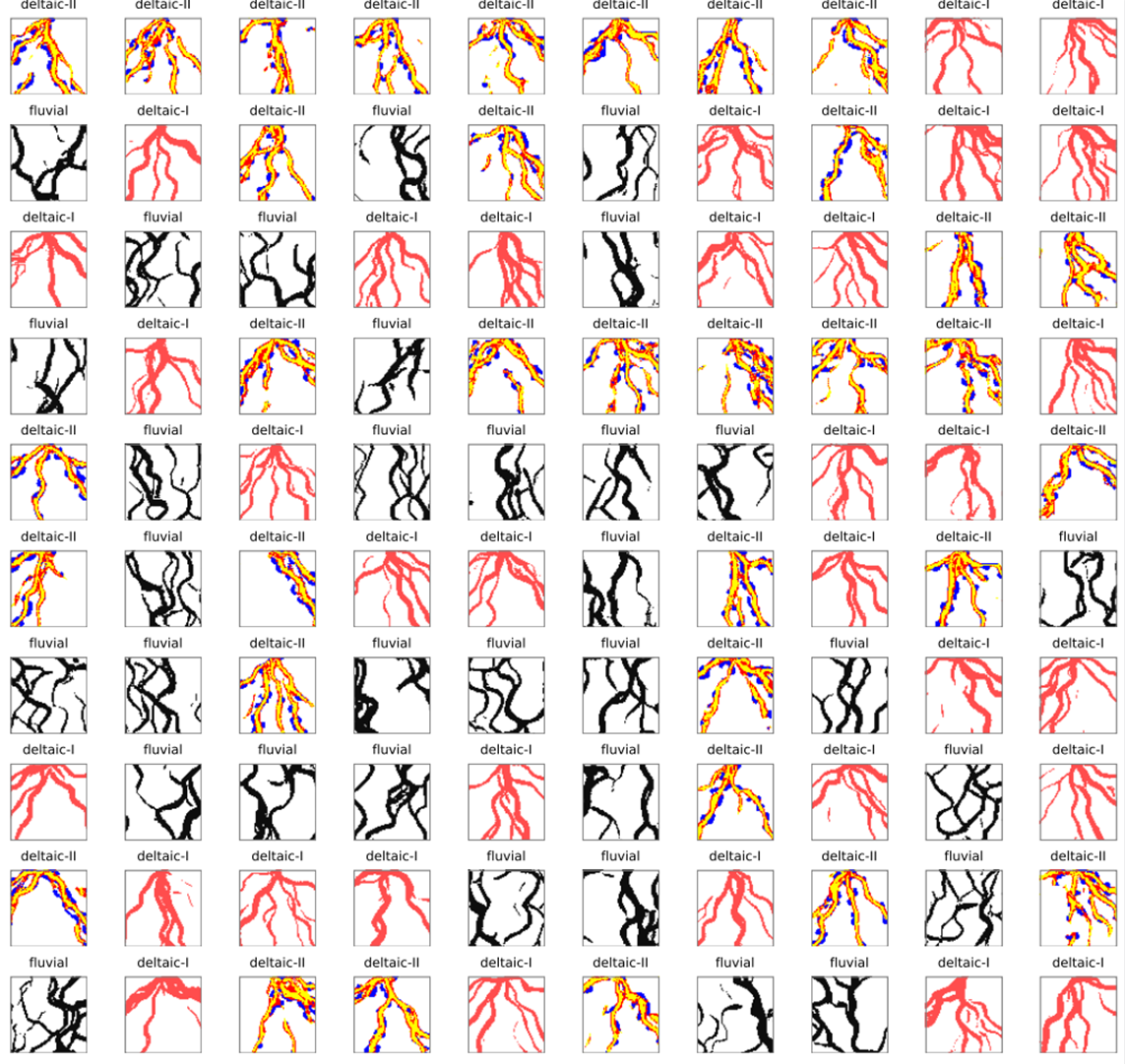}
\end{center}
\caption{100 samples by Info-WGAN show that Info-WGAN can generate a mix of depositional systems (fluvial, deltaic-I, deltaic-II) even though they have different number of facies}
\label{fig:infogan100mixsamplesof3types}    
\end{figure}

Info-WGAN with 3 latent categorical codes (labels) are used in this case study, i.e., $\mathbf{c}$=[1, 0, 0], [0, 1, 0] and [0, 0, 1]. Figure~\ref{fig:infogan100mixsamplesof3types} shows 100 samples by the generator of the Info-WAN. It demonstrates that Info-WGAN can satisfactorily generate the mixed types of sedimentary systems with the correctly predicted labels and the ratio of each type from the training dataset (1/3 each in this case study) even though the training images have different number of facies. 

\begin{figure}[H]
\begin{center}
\includegraphics[width=0.925\textwidth]{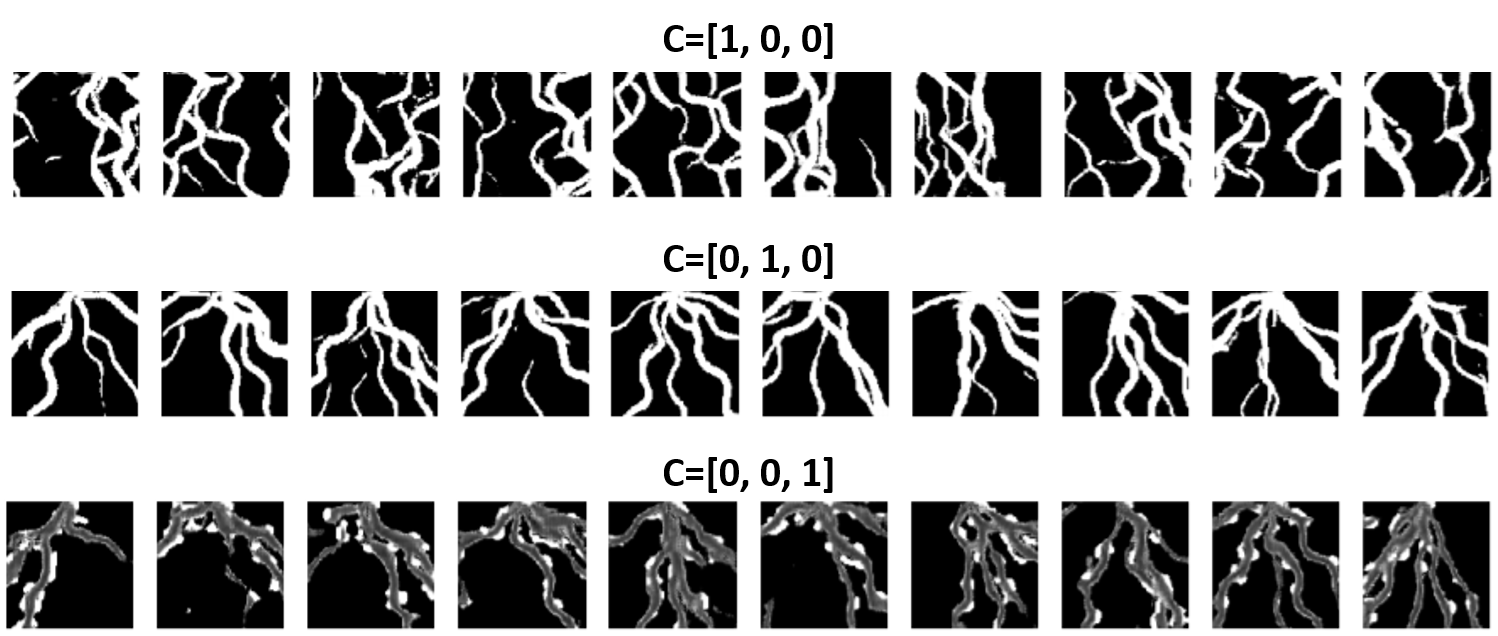}
\end{center}
\caption{The 3 types of sedimentary facies are generated by InfoGAN through disentanglement in the latent space with specified codes: binary channel system (top), binary deltic system (middle), deltic with 4 facies (bottom)}
\label{fig:codedFaciesbyInfoGAN}    
\end{figure}

Figure~\ref{fig:codedFaciesbyInfoGAN} in grayscale image shows that InfoGAN correctly generates 3 types of sedimentary systems by disentangling of the facies patterns in the latent space using maximizing infomation of 3 latent categorical codes.

This confirms the advantages of Info-WGAN as a useful tool in generating diverse samples from the training dataset by producing equal probable realizations of facies models with the correct statistics.   

\subsection{Validation on the accuracy of the predicted labels}

Another advantage of Info-WGAN over the original GANs lies in its prediction capability on the new images (facies models) since there is also a classifier $p(\mathbf{c}|\mathbf{x})$ as another output of the Info-WGAN in addition to the discriminator that tells only the probability of the generated images being real or fake. This has been illustrated in Figure~\ref{fig:infogandiagram} for the InfoGAN strcuture.

After the training of info-WGAN for the test dataset as discussed in the above section using 15000 mixed types of sedimentary systems with each type containing 5000 training images, we used additional 7000 testing images (fluvial: 2000, deltaic-I: 2000, deltaic-II: 3000), which were generated by OBM using the same statistics as those used for the training image creation, to test how good the generator of the info-WGAN can predict the labels.

Table~\ref{tbl:validationpredictionaccuracy} shows the classification accuracy matrix that tells the info-WGAN predicts the labels of the 7000 testing images with the accuracy of 99.93\% and only two fluvial images and two deltaic-I images were misclassified and all the 3000 deltaic-II images with 4 facies are correctly classified.  If the training images with partial labels are used such as in the semi-supervised training, the classification accuracy will decrease. 

%\begin{figure}[H]
%\begin{center}
%\includegraphics[width=0.90\textwidth]{Fig13.png}
%\end{center}
%\caption{Test the prediction accuracy of the labels on 3000 images that contain 2000 images for each type}
%\label{fig:gandiagram3}    
%\end{figure}

\begin{table}[H]
\begin{center}
\includegraphics[width=0.90\textwidth]{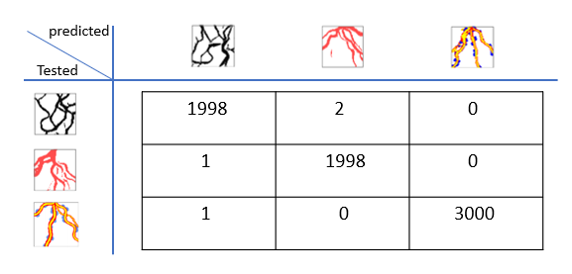}
\end{center}
\caption{Test the prediction accuracy of the labels on 7000 images that contain 2000 images for fluvial and celtaic-I repsectively and 3000 deltaic-II}
\label{tbl:validationpredictionaccuracy}    
\end{table}

\section{Conditioning the samples generated by Info-WGAN to well measurements}
\label{sec:3}

Generating geological facies models using GANs and constraining them by the well interpretations have been introduced in the paper by Dupont et al. \citep{dupont2018gangeomodeling}. The well data conditioning is done using semantic inpainting (\cite{yeh2016inpainting}; \cite{li2017inpainting}) after the training of GANs through the optimization of noisy $\mathbf{z}$-vector in the latent space by gradient descent with Adam optimization scheme. 

However, because of the mode collapse and the resulting biased sampling in the original GANs, performing data conditioning tends to be very challenging once well data locations become denser. In contrast, the Info-WGAN makes it much easier to honor dense well locations thanks to the diversity and equal probability of the samples that the Info-WGAN generated. Moreover, we have developed a novel scheme to perform the optimization of $\mathbf{z}$-vector using stochastic gradient descent by normalizing the gradient vector into a unit vector, which is a practical and useful extension of the method by Zhang et al. \citep{zhang2018hgd} about normalized direction-preserving Adam algorithm.

Our new stochastic gradient descent scheme with normalized gradient descent is written as the following:

\begin{equation}
\mathbf{v}_{t}=\beta\mathbf{v}_{(t-1)} + r\mathbf{g}_{t} 
\label{eq:5}
\end{equation}
\begin{equation}
\mathbf{z}_{t} = \mathbf{z}_{(t-1)} - \mathbf{v}_{t}
\label{eq:6}
\end{equation}

where $\mathbf{v}$ is an updated variable, $\mathbf{z}$ is the noise vector in the latent space, $r$ is the learning rate with a default value 0.006, $\mathbf{g}_{t}$ is the normalized gradient, $\beta$ is the moment factor with default value 0.999, and $t$ is the time step in the iteration process of the optimization.

To perform well data conditioning, the loss function in the optimization through error propagation over the latent noise $\mathbf{z}$-vector is designed to have two components: perceptual loss and contextual loss. While the perceptual loss penalizes unrealistic images, the contextual loss penalizes the mismatch between the generated samples and the interpreted facies at well locations \citep{dupont2018gangeomodeling} .
 
The followings will show tested cases for the well data conditioning by Info-WGAN with the new optimization scheme. It is worth noting that the data conditioning only uses the generator part of the trained Info-WGAN networks. That means, once the Info-WGAN has been trained for a training dataset, the data conditioning process can be done afterwards separately without the need to retraining the networks and this makes the generation of conditional samples by GANs very efficient, which is normally completed in seconds for one realization.

Furthermore, because of the Wasserstein distance along with the GP technique that mitigates the problem of mode collapse, it is much easier to train Info-WGANs than the original GANs. The conditioning iteration ceases once the contextual loss is below an error threshold.   

\subsection{Case 1: well data conditioning for binary fluvial facies}

\begin{figure}[H]
\begin{center}
\includegraphics[width=1.0\textwidth]{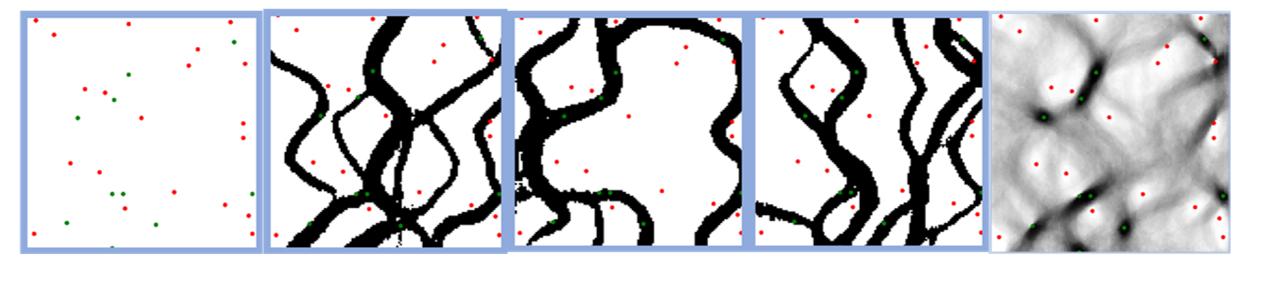}
\end{center}
\caption{Three conditional samples honoring 30 wells (left-most) by Info-WGAN and the e-type map (right-most) of 100 conditional samples; green dots are channel sand and red for shale background.}
\label{fig:infoganconditional30wellsbinary} 
\end{figure}

\begin{figure}[H]
\begin{center}
\includegraphics[width=0.95\textwidth]{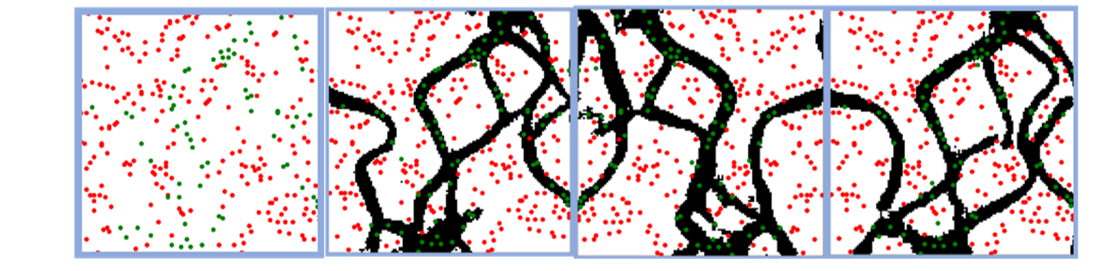}
\end{center}
\caption{Three conditional samples honoring 300 wells (left-most) by Info-WGAN}
\label{fig:infoganconditional300wellsbinary}    
\end{figure}

Figures~\ref{fig:infoganconditional30wellsbinary} and~\ref{fig:infoganconditional300wellsbinary} display conditional samples using the pre-trained Info-WGAN that honor 30 wells, 100 wells and 300 wells respectively. See Figure~\ref{fig:ganbias} for some of the 15000 binary fluvial training images (top-left). All the samples are constrained by the same set of well data and their differences indicate the uncertainty among the facies models at areas that are away from the known well locations. 

In Figure~\ref{fig:infoganconditional30wellsbinary}, the top-right map is the conditional e-type map constrained by the 30 well locations (top-left), which is computed by averaging 100 conditional samples generated by the Info-WGAN. This map provides the sand probability after knowing the well interpretation at 30 locations and it also suggests that there is less uncertainty in the area closer to well locations and channel patterns can be oberved to some extent between wells that was learnt by the generator of GANs. 

This case study demonstrates the flexibilty and capability of Info-WGAN in generating geological facies models constrained by dense well data.

\begin{figure}[H]
\begin{center}
\includegraphics[width=0.85\textwidth]{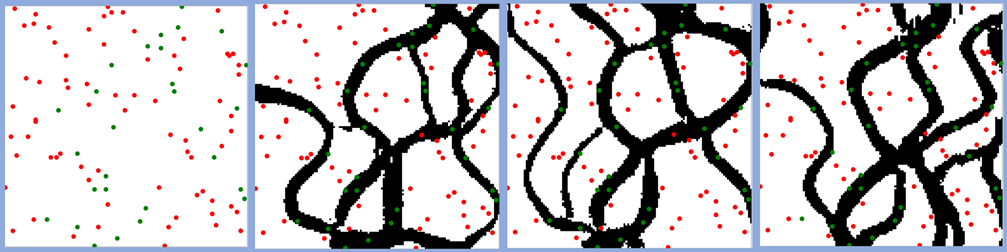}
\end{center}
\caption{Three conditional samples honoring 100 wells (left-most) by Info-WGAN}
\label{fig:infoganconditional100wellsbinary} 
\end{figure}

\subsection{Case 2: data conditioning for a mix of binary fluvial and deltaic systems}

\begin{figure} [H]
\begin{center}
\includegraphics[width=0.90\textwidth]{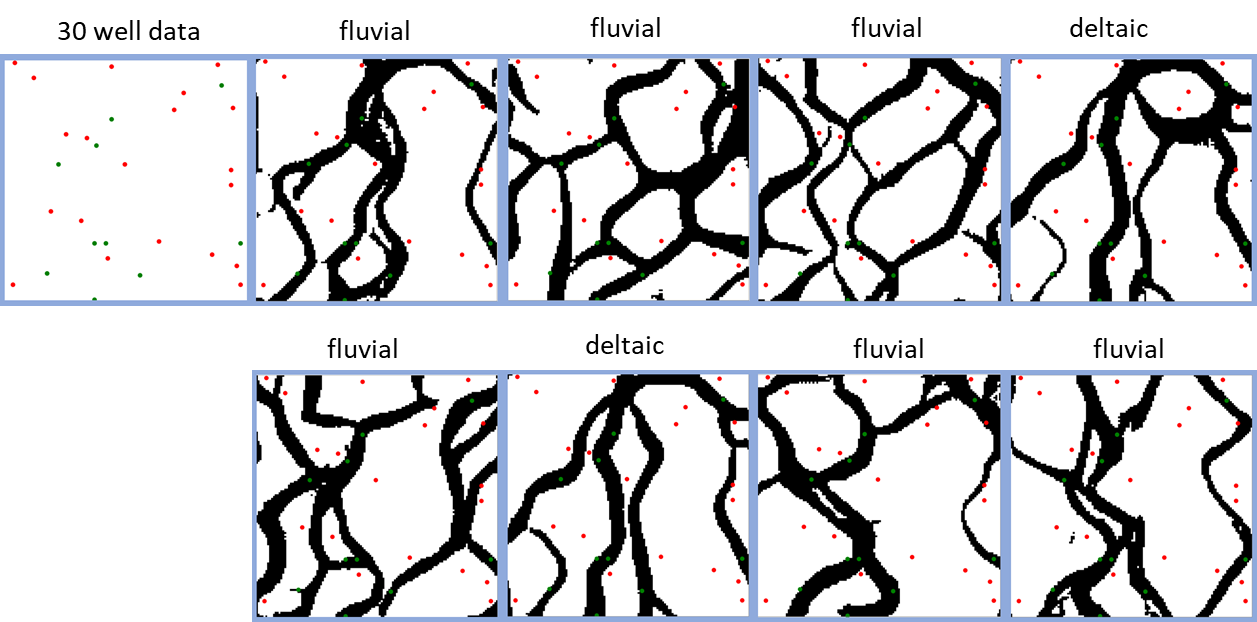}
\end{center}
\caption{30 well data locations (top-left) and 8 conditional samples by Info-WGAN. 6 of the 8 samples are fluvial deposits and the rest of 2 samples are deltaic systems and all of them honor the same set of well data}
\label{fig:infoganconditional30fluvialplusdeltaicbinary}    
\end{figure}

Figure~\ref{fig:infoganconditional30fluvialplusdeltaicbinary} demonstrates the capability of Info-WGAN in generating conditional samples when the training dataset contains mixed depositional environments such as a mix of binary fluvial and deltaic systems. The samples honor 30 wells and contain both the fluvial and deltaic deposits with the correctly predicted labels and the correct mixing ratio of the fluvial and deltaic deposits. 

This is useful, in particular for geological modeling, when the geologic senarios are uncertain and the users would like to evaluate the uncerntainty involved the variation of sedimentary facies types under the given well data constraints.
  
\subsection{Case 3: well data conditioning for multiple facies}

\begin{figure}[H]
\begin{center}
\includegraphics[width=0.80\textwidth]{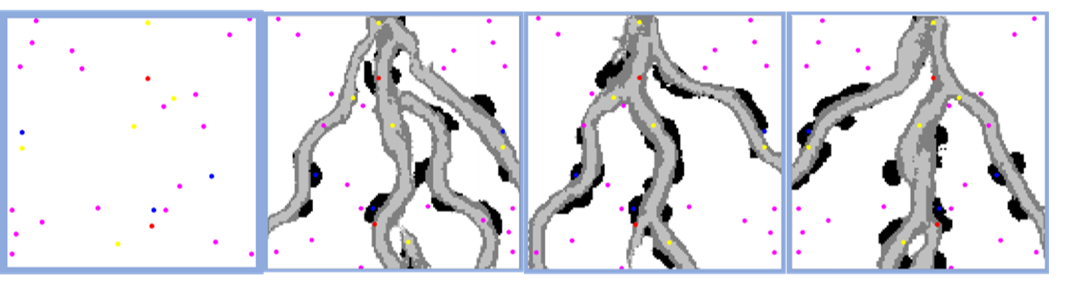}
\end{center}
\caption{Three conditional samples honoring 30 wells (left-most) by Info-WGAN using 10000 fluvial training images with 
4 facies (color legend at wells: yellow for channel, red for levee, blue for splay, magenta for shale)}
\label{fig:infoganconditional30wells4facies}    
\end{figure}

\begin{figure}[H]
\begin{center}
\includegraphics[width=1.0\textwidth]{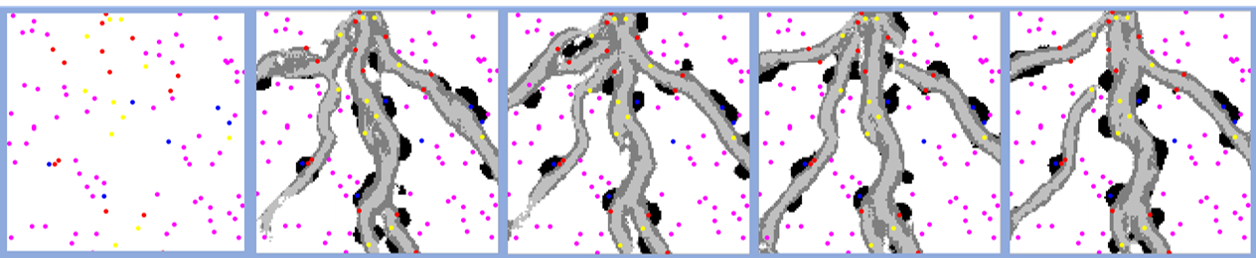}
\end{center}
\caption{Four conditional samples honoring 100 wells (left-most) by Info-WGAN using 10000 fluvial training images with 4 facies (channel, levee, splay, shale background)}
\label{fig:infoganconditional100wells4facies}    
\end{figure}

Figures~\ref{fig:infoganconditional30wells4facies} and~\ref{fig:infoganconditional100wells4facies} demonstrates the capability of Info-WGAN in generating conditional samples when the training images have multiple facies. There are 4 facies in this case study and the results show that Info-WGAN can generate realistic geology with multiple facies and condition them to various well locations. 

\section{Conclusions}
\label{sec:4}

This paper applies a novel variant of the original GANs called Info-WGAN for generating subsurface geological models constrained by well data. Compared with the original GANs, Info-WGAN can generate more diverse samples with equal probable realizations, which the original GANs often fails to provide due to the mode collapse that causes notorious difficulty in stabilizing the training of GANs. This superiority is ascribed to the disentanglement of latent variables by info-component, or infomation maximization of the method, and further boosted by Wasserstein distance and gradient descent.

By eliminating the hurdles on the diversity and ensuring a true representation of training data distribution, we believe that modeling geology using Info-WGAN is a practical and useful tool in addressing objective uncertainty and creating meaningful realizations with representative and equal probable statistics. Otherwise, the generated models by the conventional GANs would be very biased and cannot be utilized for further accurate prediction of the subsurface geology. 

The demonstrated advantages of using Info-WGAN with the aid of Wasserstein distance and gradient penalty in generating equal probable and diverse geological models would be beneficial to other deep machine learning based applications using GANs, in which more general representation and exact reproduction of the true data distribution from the training dataset become critical. The workflow and the scheme for checking the statistics can be used to determine whether the deep learning networks in image generation and modeling are representative with legitimate results. 

The following is a list of the advantages of Info-WGAN propoesed in this paper:

\begin{enumerate}
\item Applying Info-WGAN to modeling geology that combines InfoGAN with labeled geologic sedimentary types and uses Wasserstein distance and gradient penalty to overcome mode collapse of GAN training.

\item The samples generated by Info-WGAN are unbiased and as diverse as in the training images, and therefore, they can be treated as equal probable realizations.

\item Equal probable samples by Info-WGAN allow objective uncertainty evaluation, and one of them is the e-type map that is computed by averaging many generated equal probable samples to access the facies probability. These e-type maps are useful in assisting optimal decision making such as infill well drilling, reserve estimation, and the estimation of hydrocarbon flow pathways in reservoirs.

\item The latent variable used as the input to the generator network to generate new geological models can be disentangled into two parts, in which one part can have interpretable physical meaning, like $\mathbf{c}$=[0, 1] for fluvial and $\mathbf{c}$=[1, 0] for deltaic, when the mutual information maximization regularization term is included in the loss function of the Info-WGAN.

\item By adding sedimentary types as categorical codes to the latent space in addition to the noise vector $\mathbf{z}$, Info-WGAN can generate the mixed types of sedimentary environments with the correct statistics without encountering mode collapse even though the training dataset contain images with different number of facies.

\item Comparing the e-type maps between the training dataset and the samples by GANs allows to determine whether the networks are generating unbiased models, and this can be confirmed and verified by the comparison of the histograms of the e-type maps in pixels.

\item The diversity and equal probability by Info-WGAN makes the process of the well data conditioning converges faster even for much denser well locations. This fast convergence is further boosted by a novel stochastic gradient descent scheme with momentum that uses normalization of gradient vectors.
\end{enumerate}

\section*{Appendix}
\appendix
The structure of 2D Info-WGAN that generated the results for the mix of three types of sedimentary systems in Figure~\ref{fig:infogan100mixsamplesof3types} is shown in Table~\ref{tbl:infoganstructure} in Appendix. There are three networks: generator, discriminator and classifier. Both the discriminator and the classifier share the same base network that branches out into two dense layers with the size 128 towards the output layer, and then connects to two seperate outputs that provide the probability of the input image being real vs. fake and the classified lable of the input image.

We used LeakyReLU (0.2) and Dropout (0.4) in the networks. The networks were trained for 500 epochs with Adam and have a learning rate of $r$=2e-4, $\beta_1$=0.5 and $\beta_2$=0.9.The nonlinearity in the output layer of the generator is a sigmoid function.

When optimizing $\mathbf{z}$-vector in the latent space to honor the conditional data, we used the new gradient descent algorithm as described in Equations~\ref{eq:5} and~\ref{eq:6} with a learning rate of $r$=1e-6 and the momentum parameter $\beta$=0.999. We used the parameter $\lambda$=1000, a weighting factor to balance the perceptual and conextural losses, and trained for 1500 iterations for generating each conditional simulation.

\begin{table}[H]
\begin{center}
\includegraphics[width=1.0\textwidth]{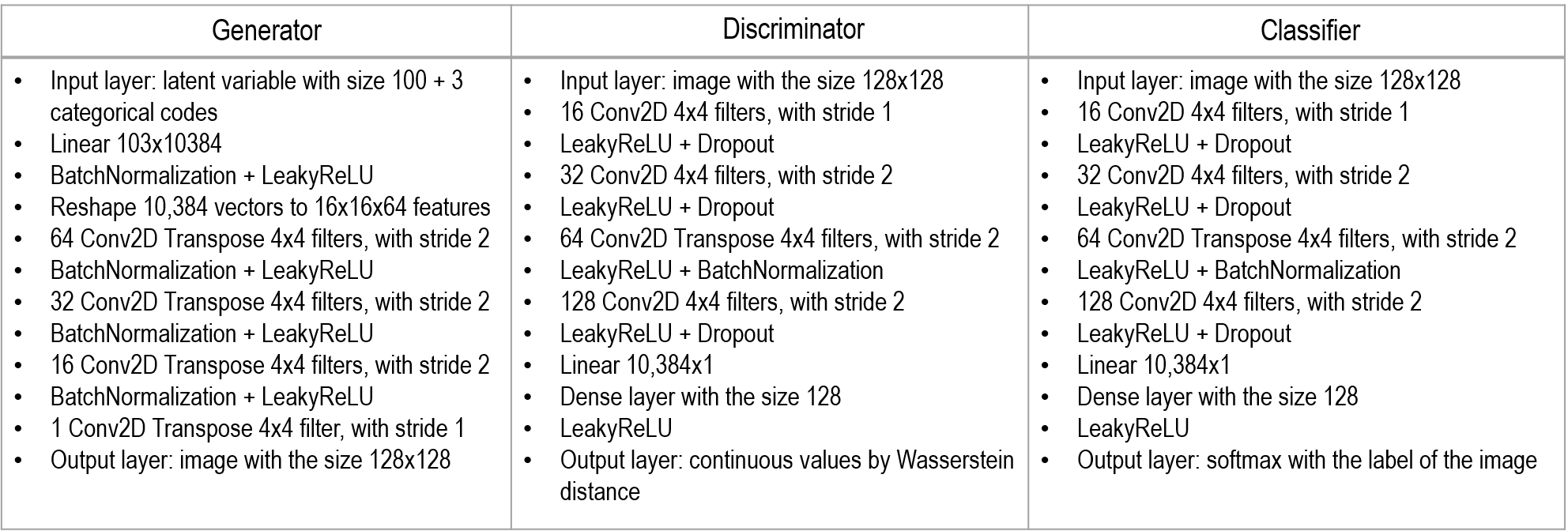}
\end{center}
\caption{Structure of the Info-WGAN networks}
\label{tbl:infoganstructure}    
\end{table}

\section*{Acknowledgement}
We would like to thank Schlumberger for the permission to publish the work, and we also thank our colleagues: Peter Tilke and Marie LeFranc for their technical discussions and support. 

%\newpage

% BibTeX users please use one of
\bibliographystyle{MG}       % Mathematical Geoscience style
{\footnotesize
\bibliography{infowganbibliography}}   % name your BibTeX data base

%% Non-BibTeX users please use
%\begin{thebibliography}{}
%%
%% and use \bibitem to create references. Consult the Instructions
%% for authors for reference list style.
%
%\bibitem{RefJ}
%% Format for Journal Reference
%Author, Article title, Journal, Volume, page numbers (year)
%% Format for books
%\bibitem{RefB}
%Author, Book title, page numbers. Publisher, place (year)
%% etc
%\end{thebibliography}

\end{document}